\documentclass[conference]{IEEEtran}
\IEEEoverridecommandlockouts
\usepackage{cite}
\usepackage{amsmath,amssymb,amsfonts}
\usepackage{algorithmic}
\usepackage{graphicx}
\usepackage{textcomp}

\usepackage{multirow}
\usepackage{times}
\usepackage{soul}
\usepackage{url}
\usepackage[hidelinks]{hyperref}
\usepackage[utf8]{inputenc}
\usepackage[small]{caption}
\usepackage{amsthm}
\usepackage{booktabs}
\usepackage{algorithm}
\usepackage{amssymb}
\usepackage[table]{xcolor}
\usepackage{xcolor}
\usepackage{svg}

\definecolor{steelblue}{rgb}{0.27,0.51,0.71}
\definecolor{cadetgrey}{rgb}{0.57,0.64,0.69}

\def\BibTeX{{\rm B\kern-.05em{\sc i\kern-.025em b}\kern-.08em
    T\kern-.1667em\lower.7ex\hbox{E}\kern-.125emX}}
\begin{document}

\newcommand{\linebreakand}{%
  \end{@IEEEauthorhalign}
  \hfill\mbox{}\par
  \mbox{}\hfill\begin{@IEEEauthorhalign}
}

\title{Characterizing Learning Dynamics of Deep Neural Networks via Complex Networks}

\author{\IEEEauthorblockN{Emanuele La Malfa\textsuperscript{1}, Gabriele La Malfa\textsuperscript{2}, Giuseppe Nicosia\textsuperscript{3} and Vito Latora\textsuperscript{4}}
\IEEEauthorblockA{
\textit{\textsuperscript{1}University of Oxford, \textsuperscript{2}University of Cambrdige, \textsuperscript{3}University of Catania, \textsuperscript{4}Queen Mary University of London}\\
\textsuperscript{1}emanuele.lamalfa@cs.ox.ac.uk, \textsuperscript{2}g.lamalfa@jbs.cam.ac.uk, \textsuperscript{3}giuseppe.nicosia@unict.it, \textsuperscript{4}v.latora@qmul.ac.uk}
% \and
% \IEEEauthorblockN{Gabriele La Malfa}
% \IEEEauthorblockA{
% \textit{{University of Cambridge}}\\
% Cambdrige, UK \\
% g.lamalfa@jbs.cam.ac.uk}
% \and
% \IEEEauthorblockN{Giuseppe Nicosia}
% \IEEEauthorblockA{
% \textit{{University of Catania}}\\
% Catania, Italy \\
% giuseppe.nicosia@unict.it
% \\gn263@cam.ac.uk}
% \centering
% \IEEEauthorblockN{Vito Latora}
% \IEEEauthorblockA{
% \textit{{Queen Mary University of London}}\\
% London, UK \\
% v.latora@qmul.ac.uk}
}

\maketitle

\begin{abstract}
In this paper, we interpret Deep Neural Networks with Complex Network Theory. Complex Network Theory (CNT) represents Deep Neural Networks (DNNs) as directed weighted graphs to study them as dynamical systems. We efficiently adapt CNT measures to examine the evolution of the learning process of DNNs with different initializations and architectures: we introduce metrics for nodes/neurons and layers, namely Nodes Strength and Layers Fluctuation. Our framework distills trends in the learning dynamics and separates low from high accurate networks. We characterize populations of neural networks (\textit{ensemble analysis}) and single instances (\textit{individual analysis}). We tackle standard problems of image recognition, for which we show that specific learning dynamics are indistinguishable when analysed through the solely Link-Weights analysis. Further, Nodes Strength and Layers Fluctuations make unprecedented behaviours emerge: accurate networks, when compared to under-trained models, show substantially divergent distributions with the greater extremity of deviations. On top of this study, we provide an efficient implementation of the CNT metrics for both Convolutional and Fully Connected Networks, to fasten the research in this direction.
\end{abstract}


\section{Introduction}
Deep Neural Networks (DNNs) have contributed in the recent years to the most remarkable progress in artificial intelligence. Algorithms reach human-comparable (or even super-human) performances in relevant tasks such as computer vision, Natural Language Processing etc.~\cite{ref_article1}. Nonetheless, it is still unclear how neural networks encode knowledge about a specific task: interpretability has thus become increasingly popular ~\cite{ref_article2} as the key to understanding enabling factors behind DNNs remarkable performances. Complex Network Theory~\cite{ref_article3} (CNT) is a branch of mathematics that investigates complex systems, from the human brain to networks of computers, by modeling and then simulating their dynamics through graphs where nodes are entities and vertices relationships \cite{chavez2010}, \cite{crucitti2004}, \cite{porta2006network}.

We address the problem of characterizing a Deep Neural Network and its training phase with (ad-hoc) CNT metrics. In this paper we formally show that this is possible and indeed insightful. As a network becomes progressively more suitable at a target task (e.g., classification of images), we analyse the dynamics of CNT metrics, to spot trends that generalize across different settings.
We focus on Fully Connected (FC) and Convolutional Neural Networks (CNNs) \cite{ref_article4} with Rectified Linear Units (\textit{ReLU}) \cite{ref_article5} activations, thus representing a step towards extending the spectrum of applications of CNT to general-purpose Deep Learning.

\section{Related Works}
Recent works have addressed the problem of analysing neural networks as directed graphs within the Complex Network Theory. In \cite{ref_scabini_cnt}, the authors propose Complex Network techniques to analyze the structure and performance of fully connected neural networks, showing high correlation to the networks classification performances. In \cite{ref_article7}, CNT metrics to distill information from Deep Belief Networks: Deep Belief Networks - which are generative models that differ from feed-forward neural networks as the learning phase is generally unsupervised - are studied with the lens of CNT by turning their architectures into a stack of equivalent Restricted Boltzmann Machines. An application of CNT metrics to feed-forward neural networks is described in \cite{ref_article8}, where the analysis is focused on the emergence of \textit{motifs}, i.e., connections that present both an interesting geometric shape and strong values of the corresponding Link-Weights. CNT has also remarkably used to assess the parallel processing capability of the network architectures \cite{petri2021topological}. These are, to the best of our knowledge, the works that model DNNs through CNT metrics. On the other hand, the evaluation of how Link-Weights (which are widely referred simply as `weights` in DNNs theory) vary during the training phase has been remarkably addressed in \cite{ref_article9}, and many other works followed \cite{ref_article10}. In view of our findings, our work connects also to the works on the dynamics - and drift - of Link-Weights \cite{shwartz2017opening}, despite our approach - which extends the analysis to `hierarchical` metrics such as Nodes Strength and Layers Fluctuation - does not involve the evaluation of quantities related to the Information Theory. From the point of view of replicability and open-sourcing code for this kind of analysis, Latora et al. \cite{ref_article3} provide an efficient implementation of a wide range of tools to study generic Complex Networks. Our work is both a theoretical and an engineering effort to translate these metrics to Deep Learning, especially when it comes to CNNs, where exploiting weights sharing and the symmetry of Convolution considerably reduces the computational time.

%%%%%%%%%%%%%%%%%%%%%%%%%%%%%%%%%%%%%%%%%%%%%%%%%%%%%%%%%%%%%%%%%%%%%%%%%%%%%%%%%%%%%%%%%%%%
% RESEARCH QUESTIONS
%%%%%%%%%%%%%%%%%%%%%%%%%%%%%%%%%%%%%%%%%%%%%%%%%%%%%%%%%%%%%%%%%%%%%%%%%%%%%%%%%%%%%%%%%%%%
\section{Contributions}
Our contribution is threefold: by extending the CNT analysis to DNNs, we answer two core research questions (RQ1, RQ2); furthermore, we provide an efficient implementation of our methods (C1) which we expect it benefits and accelerate the progress of the field. \\
\textbf{(RQ1) CNT metrics can analyse DNNs.} We introduce a formal framework to analyse Link-Weights, Nodes Strength and Layers Fluctuations of FC and Convolutional \textit{ReLU} networks. \\
\textbf{(RQ2) CNT makes the DNNs training dynamics emerge.} We train populations of networks from which we distill trends for Link-Weights, Nodes Strength and Layers Fluctuation, with the accuracy on the test set as the control parameter. Our framework is computationally efficient and can analyse CNNs with thousands of neurons on non-trivial tasks such as CIFAR10 \cite{ref_article6}. In this sense, we show that our framework is effective to characterise both populations of neural networks trained with different initializations (\textit{ensemble analysis}) as well as single instances (\textit{individual analysis}). \\
\textbf{(C1) We open-source the code to study DNNs Dynamics through CNT metrics\footnote{\label{footnote1}Download the code, along with the relative instructions and supplementary experiments from here \url{https://github.com/EmanueleLM/CNT-DNNs}}} We provide, under a permissive license, an efficient implementation of our tools to fasten the research of people interested in the topic.

%%%%%%%%%%%%%%%%%%%%%%%%%%%%%%%%%%%%%%%%%%%%%%%%%%%%%%%%%%%%%%%%%%%%%%%%%%%%%%%%%%%%%%%%%%%%
% METHODOLOGY
%%%%%%%%%%%%%%%%%%%%%%%%%%%%%%%%%%%%%%%%%%%%%%%%%%%%%%%%%%%%%%%%%%%%%%%%%%%%%%%%%%%%%%%%%%%%

\section{Methodology}
In this section we introduce the CNT notation for DNNs. We then describe the CNT metrics that we use to capture a network's learning dynamics. We finally discuss how to specifically analyse \textit{ReLU} FC and CNNs. 

\subsection{Neural Networks as Complex Networks}
W.l.o.g., we consider a supervised classification task, where an algorithm learns to assign an input vector of real numbers - $x \ \in \ \mathbb{R}^d, \ d>0, \ j \in [1,d]$  - to a categorical output from a discrete set of target classes - $c \in C$ -: to introduce the notation, we consider a generic neural network with $L>0$ hidden layers. Within each layer $\ell$, every neuron is connected through a weighted link to all the neurons in the next layer $\ell + 1$. 
The output $z_{\ell}$ of a layer $\ell$ is the product of an affine transformation in the form $z_{\ell} \ = \ \Omega_{\ell}z_{\ell-1} \ + \ \beta_{\ell}$, followed by a non-linear activation function $f_{\ell}(z_{\ell})$. The output of the last layer is a vector of real numbers $y \in \mathbb{R}^m$, from which the $argmax$ operator extracts the predicted class. We will refer to the input and output vectors respectively as $x=z_0$ and $y = z_{L}$, while $\Omega_{\ell}$ and $\beta_{\ell}$ in the previous formulae refer to the parameters of a neural network layer $\ell$. 
For an FC layer, $\Omega_{\ell}$ is a matrix of size $\mathcal{N}_{\ell}\times\mathcal{N}_{\ell+1}$ and $\beta_{\ell}$ is a vector of size $\mathcal{N}_{\ell+1}$. In a CNN layer, $\Omega_{\ell}$ is a multi-dimensional tensor employed to compute the convolution with the layer's input $z_{\ell}$. A graphic overview is sketched in Figure \ref{fig:cnngraph} (left). 

We use networks where each activation function in the hidden layers is a Rectified Linear Unit or \textit{ReLU} (i.e., $f_{\ell}(z_{\ell})=\max(0, z_{\ell})$), except for the last layer where a \textit{softmax} is applied and interpreted as the probability that the input belongs to one of the output classes, namely $f_{L}(y_{j})= exp(y_{j})/\sum_{j}^{|C|}exp(y_{j})$, being $|C|$ the cardinality of the target classes set. \\
% DNNs as CNT graphs
\indent Within the framework of CNT, a neural network is a directed bipartite graph $Net_{f}(N,E)$. Each vertex $n_{\ell,i} \ \in \ N, i \in [0, \mathcal{N}_{\ell}]$ is a neuron that belongs to a neural network layer $\ell$. The intensity of a connection - denoted as "weight" in both CNT and DNNs - is a real number assigned to the edge $(e_{n_{\ell,i}, n_{\ell+1,j}} \ \in \ E)$ that connects two neurons. Finally, $\omega^{\ell}_{i,j}$ is the weight that connects neuron $i$ from layer $\ell$ to neuron $j$ from layer $\ell+1$. 

\begin{figure}
    \centering
    \includegraphics[width=1\linewidth]{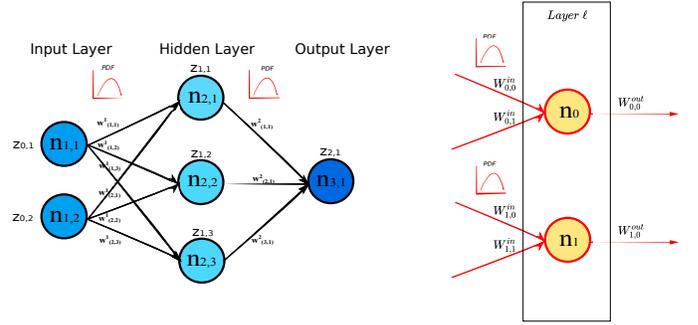}
    \caption{The Link-Weights distribution describes the parameters of a network's hidden layers (left), while Nodes Strength and Layers Fluctuation are distilled from the neurons/layers (right).}
    \label{fig:cnngraph}
\end{figure}

\subsection{Metrics for Neural Networks as Complex Networks}
We propose three CNT metrics to study the learning dynamics of DNNs. For each metric, we report a synopsis followed by an in-depth analysis (a sketch of how the metrics map to a neural network is reported is Figure \ref{fig:cnngraph}). We also report a few relevant use-cases that each metric accomplishes within our tool. \\
\textbf{Link-Weights.} The dynamic of the parameters in each layer reflects the evolution of the network training phase. \\
\textbf{Nodes Strength.} It quantifies how strong the signal is propagated through each neuron. It reveals salient flows and weak connections. \\
\textbf{Layers Fluctuation.} It extends to DNNs the notion of Nodes Fluctuation so that it is possible to measure the Strength disparity at the level of the network hidden layers. \\

\subsubsection{Link-Weights Dynamics}
As the network performs the training phase, we investigate weight dynamics in terms of mean and variance in each layer.
Given a neural network layer $\ell$, we define:

\begin{equation}
\mu^{[\ell]} = \dfrac{1}{N_{\ell} N_{\ell+1} }\sum_{i=1}^{N_{\ell}} 
\sum_{j=1}^{N_{\ell+1}} \omega^{[\ell]}_{i,j}
\end{equation}
\begin{equation}
\delta^{[\ell]} = \dfrac{1}{N_{\ell} N_{\ell+1} }\sum_{i=1}^{N_{\ell}} 
\sum_{j=1}^{N_{\ell+1}} (\omega^{[\ell]}_{i,j}-\mu^{[\ell]})^2
\end{equation}

\indent \textit{1.1) Use-cases:} The evolution of mean and variance through the training steps give significant background about learning effectiveness and stability. An underrated yet common issue is when the weight norm does not grow. This is often a symptom of model over-regularization \cite{ref_article11}. On the other hand, it is well known that where the weights grow too much, one may incur in over-fitting problems that are mitigated by regularization techniques.

\subsubsection{Nodes Strength}
The strength $s^{[\ell]}_{k}$ of a neuron $n^{[\ell]}_{k}$ is the sum of the weights of the edges incident in $n^{[\ell]}_{k}$. neural networks graphs are directed, hence there are two components that contribute to the Node Strength: the sum of the weights of outgoing edges $s^{[\ell]}_{out,k}$, and the sum of the weights of in-going links $s^{[\ell]}_{in,k}$.

\begin{equation}
s^{[\ell]}_{k} = s^{[\ell]}_{in,k} + s^{[\ell]}_{out,k} =\ \sum_{i=1}^{I}\omega^{[\ell]}_{i,k} + \sum_{j=1}^{J}\omega^{[\ell+1]}_{k,j}
\end{equation}

\indent \textit{2.1) Use-cases:}  When a neuron has an anomalously high value of Node Strength, it is propagating, compared to the other neurons in the layer, a stronger \textit{signal}: this is a hint  that either the network is not able to propagate the signal through all the neurons uniformly or that the neuron is in charge of transmitting a relevant portion of the information for the task. On the other hands, a node that propagates a weak or null \textit{signal} may be pruned (hence reducing the network complexity) as it doesn't contribute significantly to the output of the layer. A viable approach might be extreme value theory \cite{ref_article12}, as it is important to quantify the probability of a particular value to occur before deciding either to prune it or to act on the other neurons of the layer.

\subsubsection{Layers Fluctuation}
CNT identifies neural network asymmetries at nodes and links level. The standard measure, known as Nodes Disparity \cite{ref_article13}, is defined for a node $n^{[\ell]}_i$ as $Y^{[\ell]}_{i} = \sum_{i=1}^{I}[{\omega^{[\ell]}_{i}}/{s^{[\ell]}_{i}}]^2$. Nodes Disparity ranges from $0.$ to $1.$ with the maximum value when all the weights enter a single link. Conversely, weights that are evenly distributed cause the nodes in the networks to have the same - minimum - value of Disparity.
Nonetheless Disparity and similar metrics are widely adopted for studying Complex Networks \cite{ref_article14}, a fundamental problem arises when weights in the previous equation assume positive and negative values, as in the case of DNNs: the denominator can be zero for either very small values of weights or concurrently as the sum of negative and positive values that are equally balanced. In addition, it is appropriate to include a metric that measures the fluctuation of strengths in each layer, as the nodes in a DNN contribute in synergy to the identification of increasingly complex patterns. 
We propose a metric to measure the Strength fluctuations in each layer, as a proxy of the complex interactions among nodes at the same depth. \\
\indent The \textit{Layers Fluctuation} is defined as:
\begin{equation}
Y^{[\ell]} = \sqrt{\dfrac{\sum_{i=1}^{I}(s^{[\ell]}_{i} - \hat{s}^{[\ell]})^2}{I}}
\end{equation}
 
where $\hat{s}$ is computed as the average value of Nodes Strength at layer $\ell$. Please note that differently from the standard \textit{Nodes Fluctuation} introduced at the beginning of this section, the \textit{Layers Fluctuation} formula drops the dependence from each specific node $n_i$ to characterize a layer's dynamics.
The advantage of this metric is to measure disparity in a way that avoids numerical problems yet allowing to describe the networks whose weights can assume any range of values, without being restricted to positives only. \\
\indent \textit{3.1) Use-cases:} \textit{Layers Fluctuation} can be used to spot bottlenecks in a network, i.e., cases where a layer impedes the information from flowing uninterrupted through the architecture. In the experimental evaluation we show how Layers Fluctuation enables to spot interesting behaviors of a network while other metrics (included the Link-Weights and Nodes Strength) are not sufficient.

\subsection{DNNs Setup for CNT Analysis}
We take a number of precautions before setting up the experiments so that interpreting the dynamics of the CNT parameters is straightforward. The DNNs hidden layers are activated through \textit{ReLU} functions, which are universal approximators \cite{ref_article5}. We also normalize each input variable from the data between 0. and 1. before being fed into the network, so that the weights  - which can be either positive or negative - define through the \textit{ReLU} function the sign of each hidden neuron. In this way, CNT metrics capture the network dynamics and trends are easier to interpret as their sign depends on the parameters of the network. To further improve the interpretability of the results, CNT metrics can be extracted neglecting the output layer (\textit{softmax} activation) and replacing it, after the training phase, with a linear activation: we note that this operation is legitimate only for the last layer \cite{ref_article13}, where the neuron that receives the weights with the greatest magnitude is selected as the placeholder of the output class. 

\section{Experimental Results}
In this section, we provide the results of our framework. We initially describe the neural architectures that we employ, their parameters, and the data that serve as test bed. We then detail results for a population of diverse neural networks (\textit{ensemble analysis}), and we end the Section with the case of a single network analysis (\textit{individual analysis}).

\subsection{Experimental Outline}
\subsubsection{Settings} In the \textit{ensemble analysis}, parameters of FC and CNNs are sampled initially from either normal or uniform family distributions\footnote{For reasons of space, in the main paper we discuss the results for the networks whose parameters are initialized with normal distributions, while the analysis with uniform initialization is reported in Supplement, check this footnote \ref{footnote1}.} with initial variance/support being either $0.05$ or $0.5$. The topology of each network is fixed a-priori: 4 layers for the \textit{ensemble analysis} (4 dense layers for FCs, 2 Convolutional + 2 dense layers for CNNs), 6 layers for the \textit{individual analysis} (4 Convolutional + 2 dense layers). We perform separate experiments where we vary the size of the neurons/nodes in the hidden layers so that we study 2 families of models with respectively a number of parameters in the order of $10^4$ and $10^5$.
\subsubsection{Training} 
We independently train hundreds to thousands of neural networks (for each category we have introduced), which are then clustered in bins based on the accuracy of each model on the task's test set: more details on the experimental setup are provided in Table \ref{tab:data}.A. Accuracy ranges from $0.1$ (random-guess) to $1.$, with $10$ bins each of size $0.1$. Each bin contains at least $50$ networks.
Networks segmented in this way are obtained through a mix of learning and early stopping. On the other hand, identifying classes of networks whose accuracies saturate is very hard and out of the scope of this work\footnote{Finding networks that naturally stop at a specific level of accuracy is hard for the following reasons: (i) when network topology and task are fixed, it is difficult to obtain several networks that cover the whole - low/high accuracy - performance spectrum; (ii) when incorporating networks with different topologies, one incurs unsustainable computational costs and difficulties to homogenize the \textit{ensemble analysis}, especially for Nodes Strength and Layers Fluctuation where different numbers of neurons/links pose issues when comparing varying architectures.}. As a final note, while we report the results for a single round of experiments, we calculate and plot the metrics multiple times hence randomizing the composition of the populations involved so that the analysis is statistically sound.
%We use the Jensen-Shannon Divergence \cite{} as a measure of disparity between networks with different accuracy. 
\subsubsection{CNT Analysis}
We choose MNIST \cite{ref_article15} as a test bed for \textit{ensemble analysis}\footnote{Despite we could extend the \textit{ensemble analysis} to increasingly complex tasks (e.g. CIFAR10) without further adjustments to the framework, we leave it as future work. This non-trivial extension is beyond the scope of this work and involves a massive amount of computational resources.}. We normalize the data-sets before training so that each input dimension spans from $0.$ to $1.$, while DNNs weights are unbounded in $\mathbb{R}$.
As a complement to the \textit{ensemble analysis} and as an extension of the work in \cite{ref_article8}, we use our framework to study the dynamics of individual DNNs on CIFAR10 \cite{ref_article6}, where we compute CNT metrics for different snapshots (at different levels of accuracy) of a single neural network. Patterns that are specific to that instance are hence \textit{local}. The details on both the \textit{ensemble} and \textit{local} approaches are reported in Table \ref{tab:data}.B.

\subsubsection{Results Visualization} In the \textit{ensemble} analysis, we compute and plot the CNT metrics as Probability Distribution Functions (PDF) or alternatively as error bars. In each Figure, the population of lowest-performing networks is plotted in \textcolor{cadetgrey}{grey}, while the best performing network's PDF is plotted in \textcolor{blue}{blue}: we use color scale (from grey to blue) to plot increasingly performing networks. We note that the same approach is applied to plot the dynamics in the \textit{individual} analysis. 

\subsection{Results}
\subsubsection{Trends in \textit{Ensemble Analysis}}
When comparing low vs. high performing FC networks trained on MNIST, Link-Weights analysis distills patterns that only partially capture the complexity of the training dynamics, as evidenced in Figure \ref{fig:LW_05_005_Normal_Small}. In fact, low-performing networks are hardly distinguishable from their high-performing correspondents. We note that the Link-Weights PDFs tend to flatten out slightly as performances improve, generating heavy tails with stable mean values. The absence of a strong \textit{discrimen} between low and high performing networks becomes even more severe when analysing CNNs (Figure \ref{fig:CNN_LW_05_005_Normal_Small}, top), thus providing motivation in pursuing an analysis that encompasses increasingly complex metrics such as Nodes Strength and Layers Fluctuations.  
The initial distribution of the Nodes Strength, which in the early stages of learning is characterized by zero-mean and a small variance, tends to widen its support as the train increases the network's accuracy. As sketched in Figure \ref{fig:NS_05_005_Normal_Small}, high-performing FC networks with support $\pm 0.05$, exhibit values of Nodes Strength that concentrate around positive or negative values, relegating a contribution close to zero to a restricted number of nodes. Interestingly, in a few hidden layers the Nodes Strength PDFs is bimodal (Figure \ref{fig:NS_05_005_Normal_Small} bottom, layers 2 and 3) while in general heavy tails characterize accurate networks. 
As the variance of the Nodes Strength increases, Layers Fluctuation tends to flat with the PDF that in a few cases becomes multi-modal as model accuracy increases (Figure \ref{fig:LF_05_005_Normal_Small}).  

For FC networks initialized with Normal distributions and support 0.5, we report results from a batch of experiments with dynamics that are worth discussing: the networks with highest accuracy (Figures \ref{fig:LW_05_005_Normal_Small}, \ref{fig:NS_05_005_Normal_Small}, \ref{fig:LF_05_005_Normal_Small}, top row, \textcolor{blue}{blue} PDFs) significantly deviate from the behavior of the precedents as they exhibit restrained variance. Similar observations - trhough targeting different research questions - have been made in Shwartz-Ziv et al. \cite{shwartz2017opening}, where they analyse mean and variance of a network's weights at different phases of learning, with the training epochs as control parameter (the reader can refer to Figures 4 and 8 in \cite{shwartz2017opening}). While so far patterns in the learning dynamics have been observed only for single neural instances, our approach permits to appreciate these at a global level.

The same analysis conducted on a population of CNNs evidences similar trends for all the three metrics we evaluate, as sketched in Figures \ref{fig:CNN_LW_05_005_Normal_Small}, \ref{fig:CNN_NS_05_005_Normal_Small}, \ref{fig:CNN_LF_05_005_Normal_Small}. 
%\GLM{The first two layers are more noised probably due to the presence of a low number of parameters.}
Highly discriminant, monotonic behaviors of the PDFs are especially evident when analysing the Nodes Strength, where the Convolutional layers (Figure \ref{fig:CNN_NS_05_005_Normal_Small}, the first 3 plots from the respective rows) are characterized by \textit{fat positive tails}, a hint that in a population of networks initialized with that configuration, which we remember is the same for all the networks, very different yet highly accurate models co-exist.
We complement the qualitative analysis exposed so far, with the PDFs statistical quantification of skewness and kurtosis (Figures \ref{fig:Stat_NS_LF_05_005_Normal_Small} and \ref{fig:Stat_NS_LF_05_005_Normal_Small_CNN}).
%\ELM{Maybe we can discuss this more} 

%From this analysis, we can conclude that accurate FC and CNNs have higher probability to exhibit Nodes Strengths and Layers Fluctuations that are non-homogeneous, thus evidencing paths where information flows more easily than in others. 
%We sketch this trends in Figure \ref{fig:fc-beyond-weights}. 

\subsubsection{Trends in \textit{Individual Analysis}}
We train a $6$ layers CNN - 4 Convolutional + 2 dense layers - on CIFAR10, reaching different levels of accuracy - up to $0.7$ - and we show that: (i) CNT is both a \textit{local} and a \textit{global} method, i.e., it shows trends in the training dynamics for both populations and individual neural networks; (ii) it is not necessary to have models with strong performances to spot salient behaviors. We observe that, in continuity with the \textit{ensemble analysis}, the Nodes Strength of Convolutional layers monotonically decreases (Figure \ref{fig:LOCAL_CNN}, second row). For accurate models, in the PDFs of the Convolutional layers, we observe the emergence of numerous spikes, a hint that specific input pixels strongly activate a few network's receptive fields, while the majority of the weights is less stimulated. Notably, FC layers do not show this \textit{motif} but tend to assume more smooth and regular shapes (see previous discussion and Figures \ref{fig:CNN_NS_05_005_Normal_Small} and \ref{fig:NS_05_005_Normal_Small}).
As regards the \textit{individual analysis} of Layers Fluctuation (Figure \ref{fig:LOCAL_CNN}, bottom row), we provide an alternative (to the PDFs) yet equivalent visualization method based on error-bars. Error bars are useful to visualize Layers Fluctuations when fewer data points are available, as in the case of the \textit{individual analysis} where fewer instances (of the same model) can be trained.

With the model that reaches its maximum level of accuracy, Layers Fluctuation grows monotonically, with the exception of the fifth layer, which connects the second dense layer to the third. The result is consistent with what we observe in the \textit{ensemble analysis}, thus confirming that across different architectures and tasks, accurate networks generally exhibit stronger Layers Fluctuation.  

%figure CNN
\begin{figure}
    \centering
    \includegraphics[width=1\linewidth]{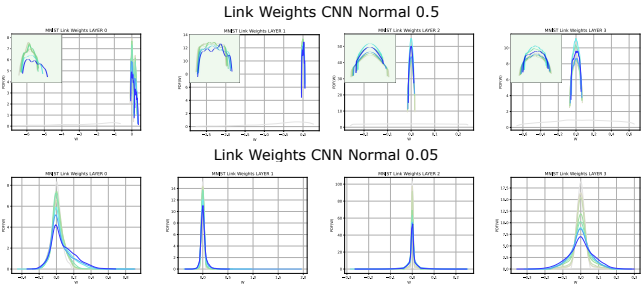}
    \caption{The figure shows the dynamics of the Link-Weights for CNNs with a Normal distribution initialization and a scaling factor respectively of 0.5 (plots on top) and 0.05 (plots on bottom). The networks are trained on MNIST. The colour scale represents the spectrum of networks at different training accuracies, from the lowest in grey to the highest in blue.}
    \label{fig:CNN_LW_05_005_Normal_Small}
\end{figure}

\begin{figure}
    \centering
    \includegraphics[width=1\linewidth]{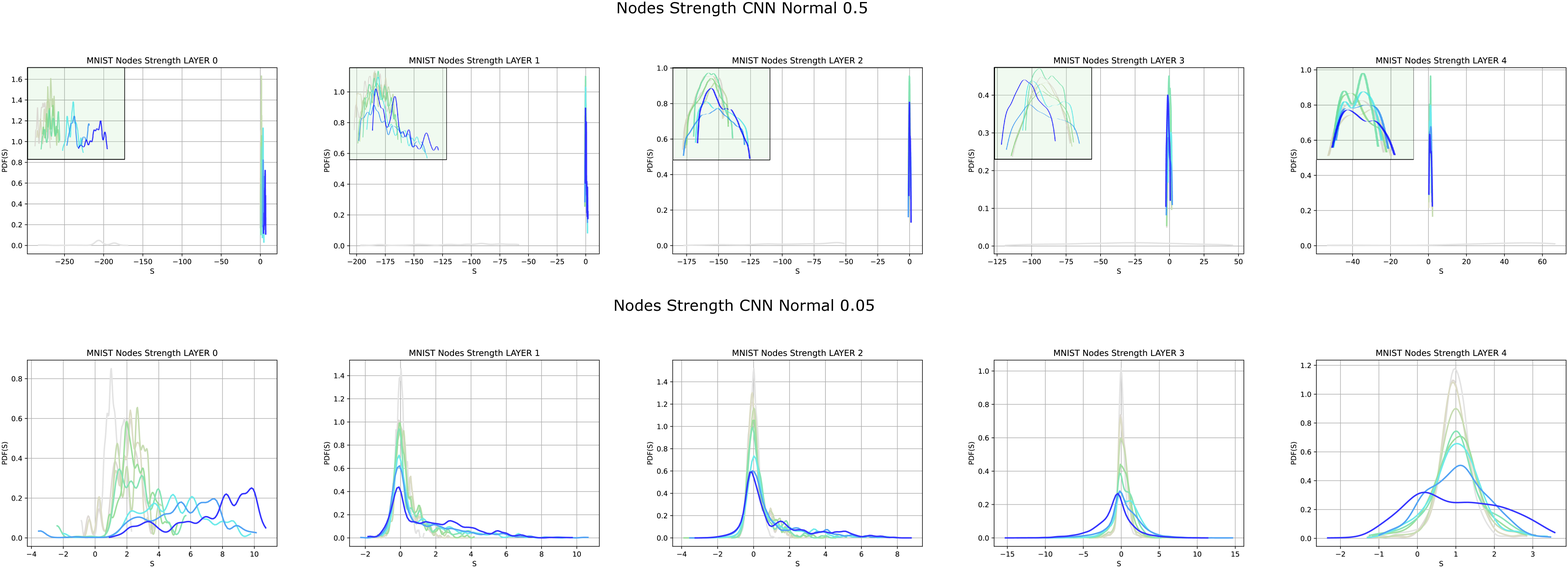}
    \caption{The figure shows the dynamics of the Nodes Strength for CNNs with a Normal distribution initialization and a scaling factor respectively of 0.5 (plots on top) and 0.05 (plots on bottom). The networks are trained on MNIST. The colour scale represents the spectrum of networks at different training accuracies, from the lowest in grey to the highest in blue.}
    \label{fig:CNN_NS_05_005_Normal_Small}
\end{figure}

\begin{figure}
    \centering
    \includegraphics[width=1\linewidth]{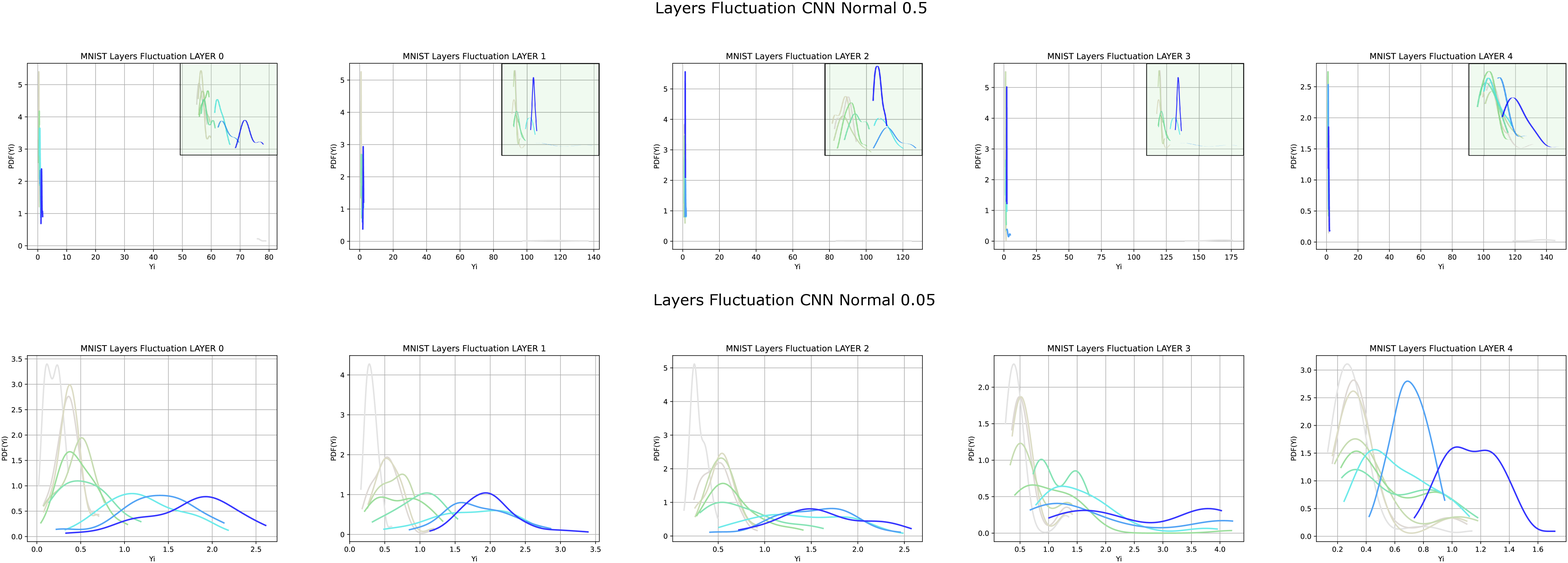}
    \caption{The figure shows the dynamics of the Layers Fluctuation for CNNs with a Normal distribution initialization and a scaling factor respectively of 0.5 (plots on top) and 0.05 (plots on bottom). The networks are trained on MNIST. The colour scale represents the spectrum of networks at different training accuracies, from the lowest in grey to the highest in blue.}
    \label{fig:CNN_LF_05_005_Normal_Small}
\end{figure}

%stats CNN
\begin{figure}
    \centering
    \includegraphics[width=1\linewidth]{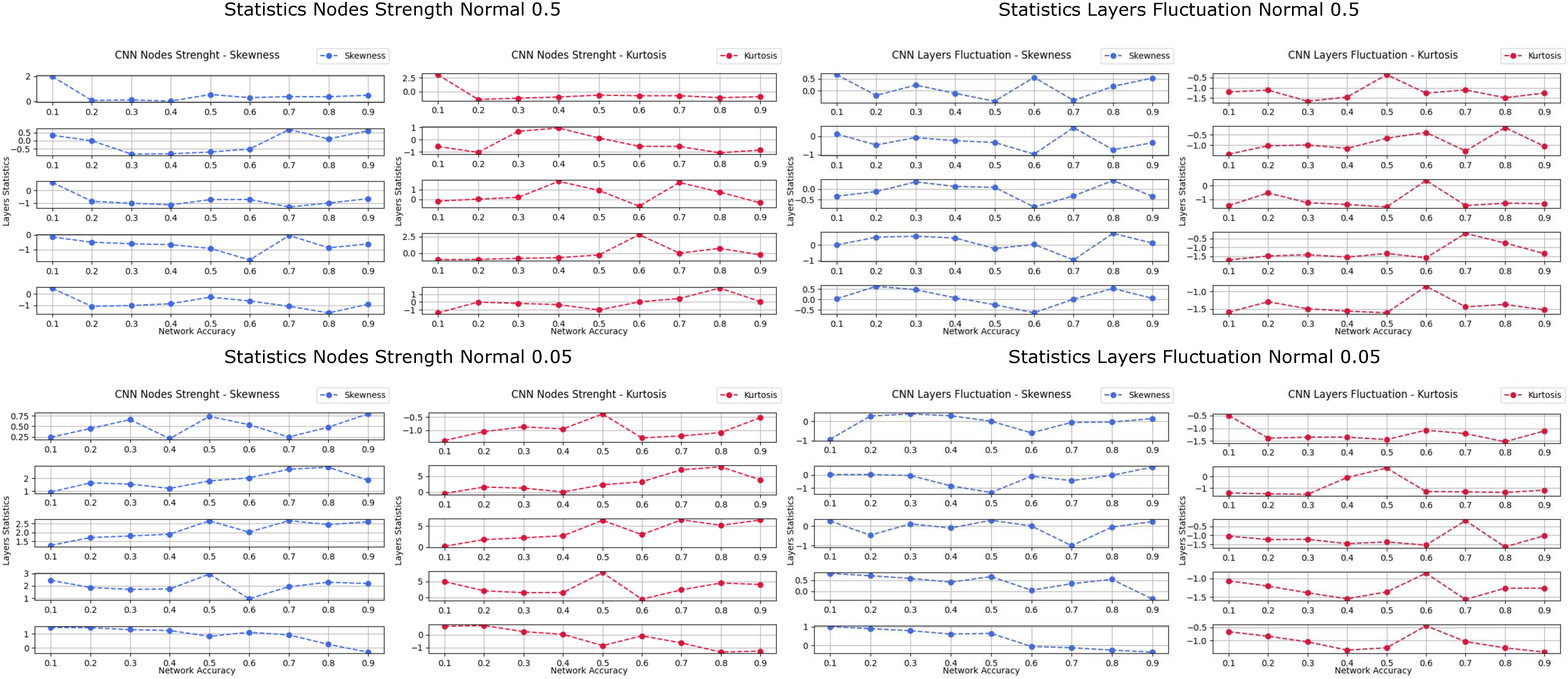}
    \caption{The figure shows the values of Kurtosis (red) and Skewness (blue) of the Nodes Strength and Layers Fluctuations for each layer, for different levels of accuracy. Values refer to CNNs with a Normal distribution initialization and a scaling factor respectively of 0.5 (left columns) and 0.05 (right columns). The networks are trained on MNIST.}
    \label{fig:Stat_NS_LF_05_005_Normal_Small_CNN}
\end{figure}

%stats CNN LOCAL
\begin{figure}
    \centering
    \includegraphics[width=1\linewidth]{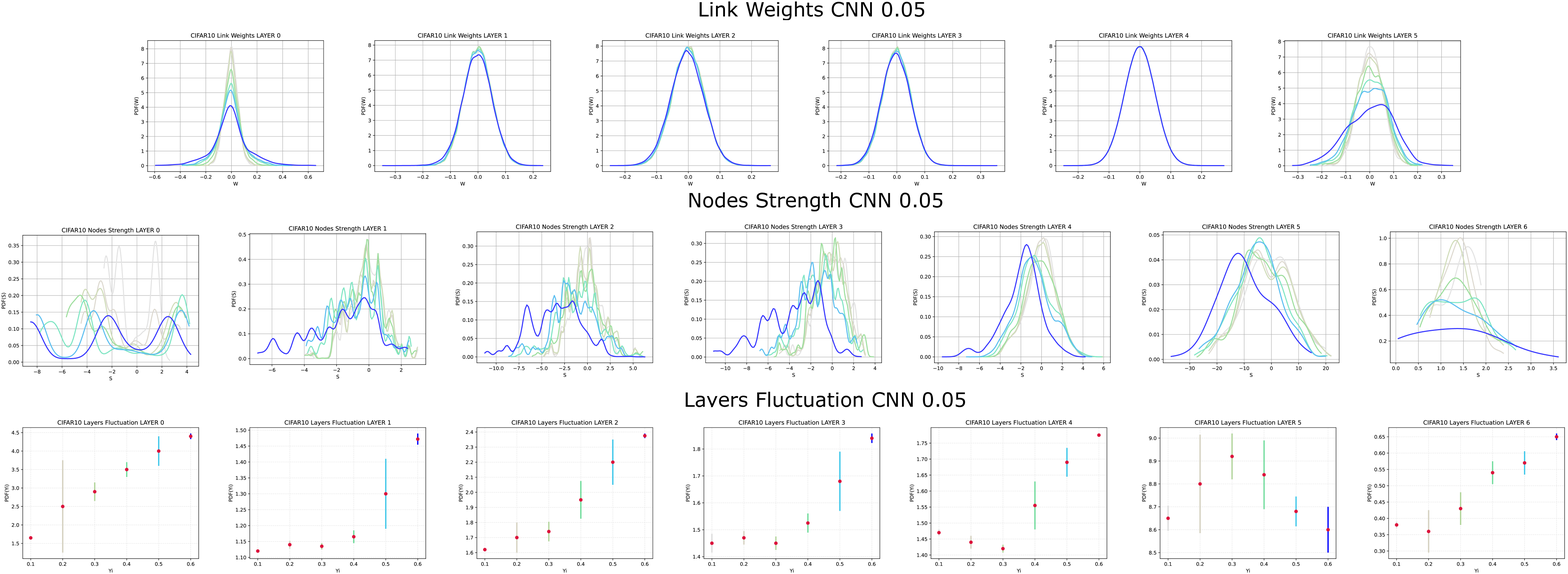}
    \caption{The figure shows the local dynamics of the Link-Weights (top), Nodes Strength (medium) and Layers Fluctuation (bottom) of CNNs initialized with Normal distribution and scaling factor of 0.05. The networks are trained on CIFAR10. The colour scale represents the spectrum of networks at different training accuracies (max accuracy reached is approximately 0.7), from the lowest in grey to the highest in blue. Please note that for the Layers Fluctuations, being this a local analysis on a restricted number of samples, we preferred to show the error bars, which are equivalent yet easier to interpret compared to the PDFs.}
    \label{fig:LOCAL_CNN}
\end{figure}

\section{Conclusions}
We have presented a new framework to interpret DNNs using CNT. We study the learning phase of Fully Connected (FC) and Convolutional Neural Networks (CNN) leveraging three metrics: Link-Weights, Nodes Strength and Layers Fluctuation. Nodes Strength and Layers Fluctuation are ad-hoc metrics that complement the Link-Weights analysis, evidencing unprecedented behaviours of the networks. With CNT, we study two complementary approaches: the former on a population of networks initialized with different strategies (\textit{ensemble analysis}), the latter on different snapshots of the same instance (\textit{individual analysis}). As FCs and CNNs get accurate on the learning task, the Probability Density Functions (PDFs) of Nodes Strength and Layers Fluctuation gradually flatten out. This analysis allows to discriminate between low and high performing networks and their learning phases.

As a future work, we will extend the CNT framework to different and/or more complex architectures, such as Recurrent Neural Networks and Attention-based models. Another viable extension would encompass diverse tasks such as NLP and signal processing.

%figure Fully Connected
\begin{figure*}
    \centering
    \includegraphics[width=1\linewidth]{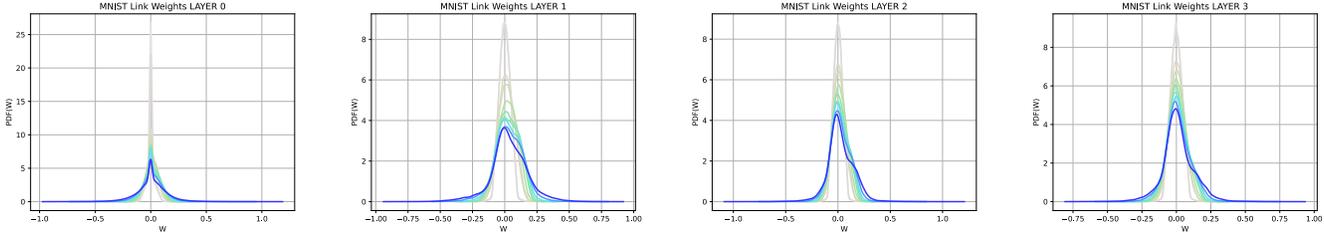}
    \caption{The figure shows the dynamics of the Link-Weights for FC networks with a Normal distribution initialization and a scaling factor respectively of 0.5 (plots on top) and 0.05 (plots on bottom). The networks are trained on MNIST. The colour scale represents the spectrum of networks at different training accuracies, from the lowest in grey to the highest in blue.}
    \label{fig:LW_05_005_Normal_Small}
\end{figure*}

\begin{figure*}
    \centering
    \includegraphics[width=1\linewidth]{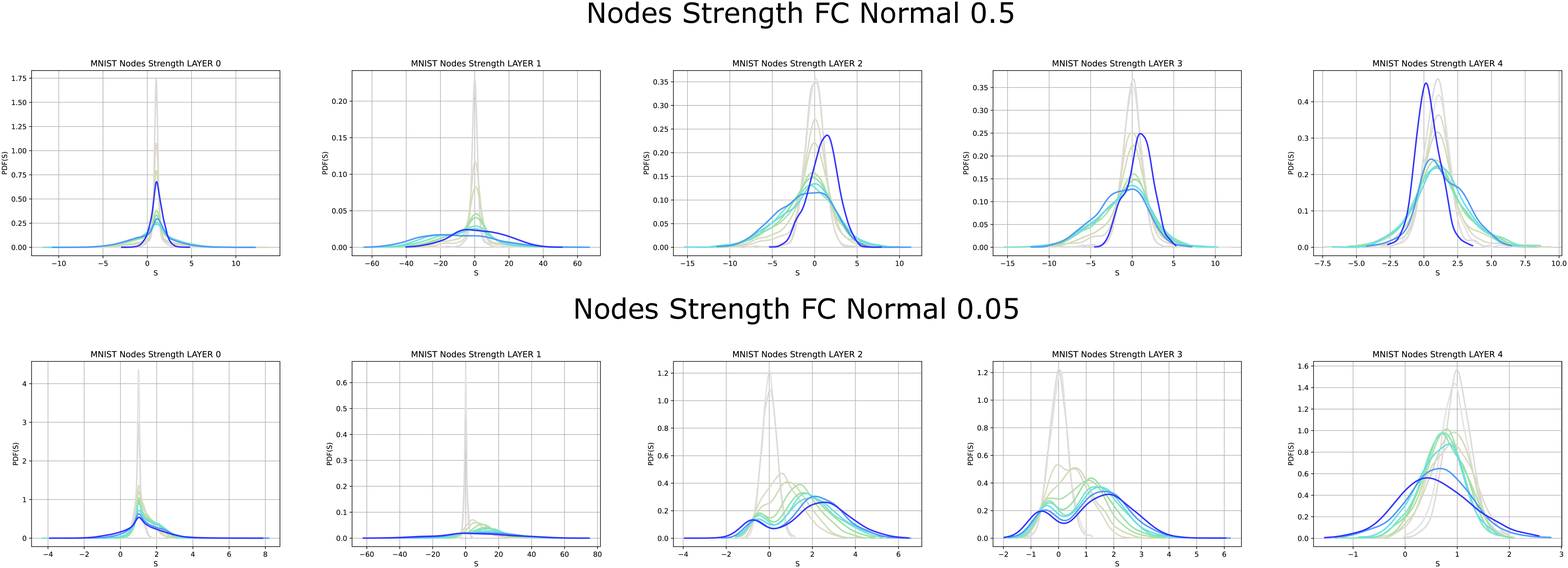}
    \caption{The figure shows the dynamics of the Nodes strength for FC networks with a Normal distribution initialization and a scaling factor respectively of 0.5 (plots on top) and 0.05 (plots on bottom). The networks are trained on MNIST. The colour scale represents the spectrum of networks at different training accuracies, from the lowest in grey to the highest in blue.}
    \label{fig:NS_05_005_Normal_Small}
\end{figure*}

\begin{figure*}
    \centering
    \includegraphics[width=1\linewidth]{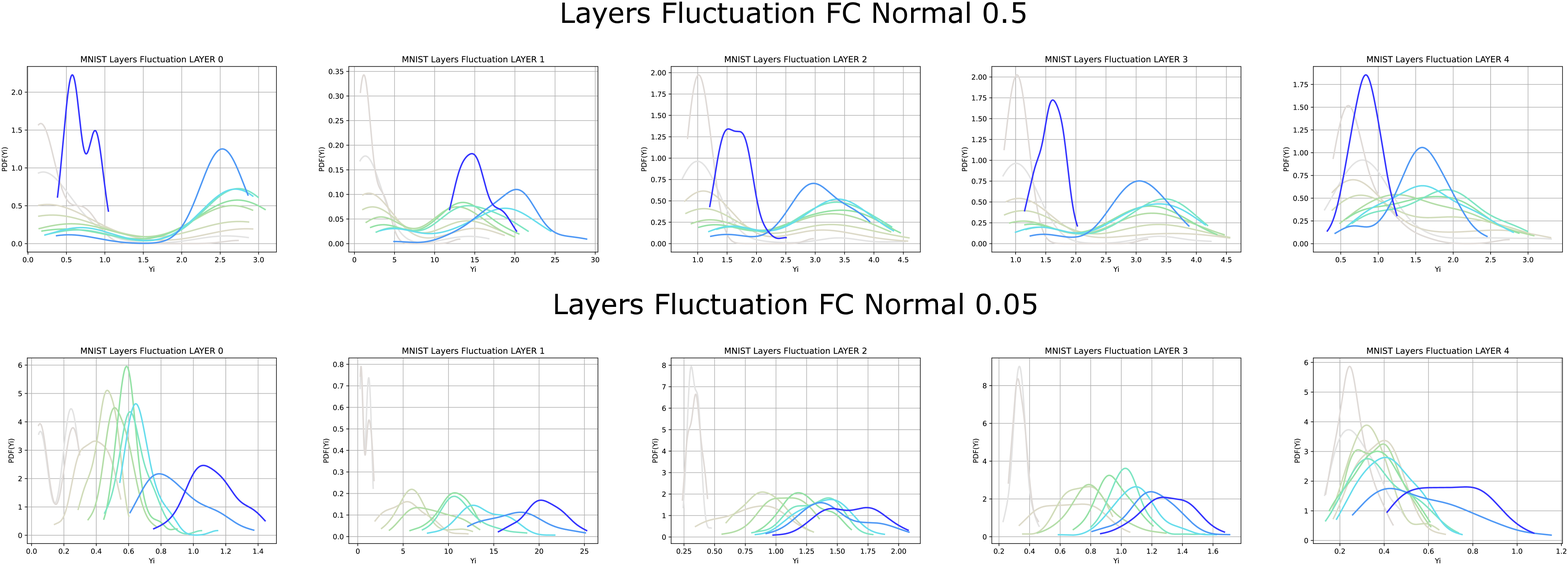}
    \caption{The figure shows the dynamics of the Layers Fluctuation for FC networks with a Normal distribution initialization and a scaling factor respectively of 0.5 (plots on top) and 0.05 (plots on bottom). The networks are trained on MNIST. The colour scale represents the spectrum of networks at different training accuracies, from the lowest in grey to the highest in blue.}
    \label{fig:LF_05_005_Normal_Small}
\end{figure*}

%stats FC
\begin{figure*}
    \centering
    \includegraphics[width=1\linewidth]{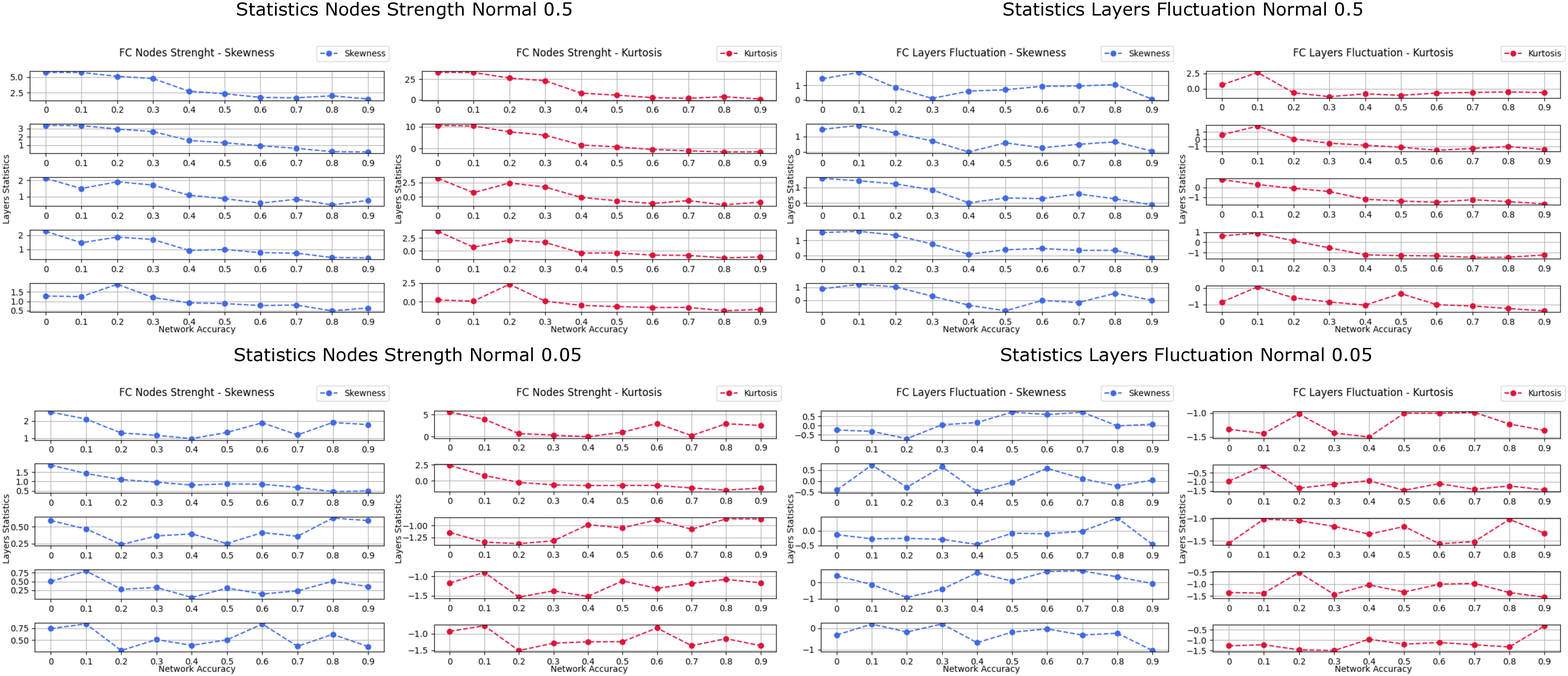}
    \caption{The figure shows the values of Kurtosis (red) and Skewness (blue) of the Nodes Strength and Layers Fluctuations for each layer, for different levels of accuracy. Values refer to FC networks with a Normal distribution initialization and a scaling factor respectively of 0.5 (top) and 0.05 (bottom). The networks are trained on MNIST.}
    \label{fig:Stat_NS_LF_05_005_Normal_Small}
\end{figure*}

\begin{table*}
\centering
\begin{tabular}{|c|c|c|c|c|c|c|}
\hline
 \multicolumn{7}{|c|}{\textbf{Table I.A: Ensemble Analysis - MNIST Dataset}}\\
 \hline
 \hline
 \textbf{Analysis} &  \small{\textbf{Network}} & \small{\textbf{Max Accuracy}} & \small{\textbf{Layers}} & \small{\textbf{Scaling Factor}} & \small{\textbf{Init Methods}} &  \small{\textbf{figure*}}\\ 
 \hline
\multirow{1}{*}{\small{Link-Weights}} &  \small{FC} & \small{$0.98$} &  \small{$4$} &  \small{0.5 - 0.05} &  \small{normal} &  \small{\ref{fig:LW_05_005_Normal_Small}}\\ \cline{1-7}
  
\multirow{1}{*}{\small{Strength}}&{\small{FC}} & \small{$0.98$} &  \small{$4$} &  {\small{0.5 - 0.05}} &  \small{normal} &  \small{\ref{fig:NS_05_005_Normal_Small}} \\ \cline{1-7}

\multirow{1}{*}{\small{Fluctuation}}&{\small{FC}} & \small{$0.98$} &  \small{$4$} &  {\small{0.5 - 0.05}} &  \small{normal} &  \small{\ref{fig:LF_05_005_Normal_Small}} \\ \cline{1-7}

\multirow{1}{*}{\small{Link-Weights}} &  \small{CNN} & \small{$0.98$} &  \small{$4$} &  \small{0.5 - 0.05} &  \small{uniform} &  \small{\ref{fig:CNN_LW_05_005_Normal_Small}}\\ \cline{1-7}

\multirow{1}{*}{\small{Strength}}&{\small{CNN}} & \small{$0.98$} &  \small{$4$} &  {\small{0.5 - 0.05}} &  \small{uniform} &  \small{\ref{fig:CNN_NS_05_005_Normal_Small}} \\ \cline{1-7}

\multirow{1}{*}{\small{Fluctuation}}&{\small{CNN}} & \small{$0.98$} &  \small{$4$} &  {\small{0.5 - 0.05}} &  \small{uniform} &  \small{\ref{fig:CNN_LF_05_005_Normal_Small}} \\ \cline{1-7}

 \hline 
 \hline
 \multicolumn{7}{|c|}{\textbf{Table I.B: Individual Analysis - CIFAR10 Dataset}}\\
 \hline
 \hline
 \textbf{Analysis} &  \small{\textbf{Network}} & \small{\textbf{Max Accuracy}} & \small{\textbf{Layers}} &  \small{\textbf{Scaling Factor}} & \small{\textbf{Init Methods}} &  \small{\textbf{figure*}}\\ 
 \hline
 
  \multirow{1}{*}{\small{Link-Weights}}&{\small{CNN}} & \small{$0.7$} &  \small{$6$} &  {\small{0.05}} &  \small{normal} &  \small{\ref{fig:LOCAL_CNN}}\\ \cline{1-7}

 \multirow{1}{*}{\small{Strength}}&{\small{CNN}} & \small{$0.7$} &  \small{$6$} &  {\small{0.05}} &  \small{normal} &  \small{\ref{fig:LOCAL_CNN}}\\ \cline{1-7}

 \multirow{1}{*}{\small{Fluctuation}}&{\small{CNN}} & \small{$0.7$} &  \small{$6$} &  {\small{0.05}} &  \small{normal} &  \small{\ref{fig:LOCAL_CNN}}\\ \cline{1-7}

\end{tabular}
\caption{The table reports the models' characteristics of the \textit{ensemble analysis} and the \textit{individual analysis} presented in the paper. We provide the type of network, the parameters, and the test data-sets. Each experiment is referenced to the respective Figure(s)in the paper.}
\label{tab:data}
\end{table*}

%
% ---- Bibliography ----
%
% BibTeX users should specify bibliography style 'splncs04'.
% References will then be sorted and formatted in the correct style.
%
% \bibliographystyle{splncs04}
% \bibliography{mybibliography}
%

\end{document}